\PassOptionsToPackage{table,xcdraw}{xcolor}

\documentclass[10pt,twocolumn,letterpaper]{article}

\usepackage{graphicx}
\usepackage{multirow}

\usepackage[pagenumbers]{cvpr} % To force page numbers, e.g. for an arXiv version

\definecolor{cvprblue}{rgb}{0.21,0.49,0.74}
\usepackage[pagebackref,breaklinks,colorlinks,allcolors=cvprblue]{hyperref}

\def\paperID{13947} % *** Enter the Paper ID here
\def\confName{CVPR}
\def\confYear{2025}

\title{MEGL: Multimodal Explanation-Guided Learning}

\author{
Yifei Zhang\thanks{Equal contribution.}\\
Emory University\\
{\tt\small yifei.zhang2@emory.edu}
\and
Tianxu Jiang\footnotemark[1]\\
University of Michigan\\
{\tt\small tianxuj@umich.edu}
\and
Bo Pan\\
Emory University\\
{\tt\small bo.pan@emory.edu}
\and
Jingyu Wang\\
Emory University\\
{\tt\small jingyu.wang@emory.edu}
\and
Guangji Bai\\
Emory University\\
{\tt\small guangji.bai@emory.edu}
\and
Liang Zhao\\
Emory University\\
{\tt\small liang.zhao@emory.edu}
}

\begin{document}
\maketitle
\begin{abstract}
Explaining the decision-making processes of Artificial Intelligence (AI) models is crucial for addressing their “black box” nature, particularly in tasks like image classification. Traditional eXplainable AI (XAI) methods typically rely on unimodal explanations, either visual or textual, each with inherent limitations. Visual explanations highlight key regions but often lack rationale, while textual explanations provide context without spatial grounding. Further, both explanation types can be inconsistent or incomplete, limiting their reliability. To address these challenges, we propose a novel Multimodal Explanation-Guided Learning (MEGL) framework that leverages both visual and textual explanations to enhance model interpretability and improve classification performance. Our Saliency-Driven Textual Grounding (SDTG) approach integrates spatial information from visual explanations into textual rationales, providing spatially grounded and contextually rich explanations. Additionally, we introduce Textual Supervision on Visual Explanations to align visual explanations with textual rationales, even in cases where ground truth visual annotations are missing. A Visual Explanation Distribution Consistency loss further reinforces visual coherence by aligning the generated visual explanations with dataset-level patterns, enabling the model to effectively learn from incomplete multimodal supervision. We validate MEGL on two new datasets, Object-ME and Action-ME, for image classification with multimodal explanations. Experimental results demonstrate that MEGL outperforms previous approaches in prediction accuracy and explanation quality across both visual and textual domains. Our code will be made available upon the acceptance of the paper.
\end{abstract}    
\section{Introduction}

\label{sec:intro}
\begin{figure}[htbp]
\centering
\includegraphics[width=\linewidth]{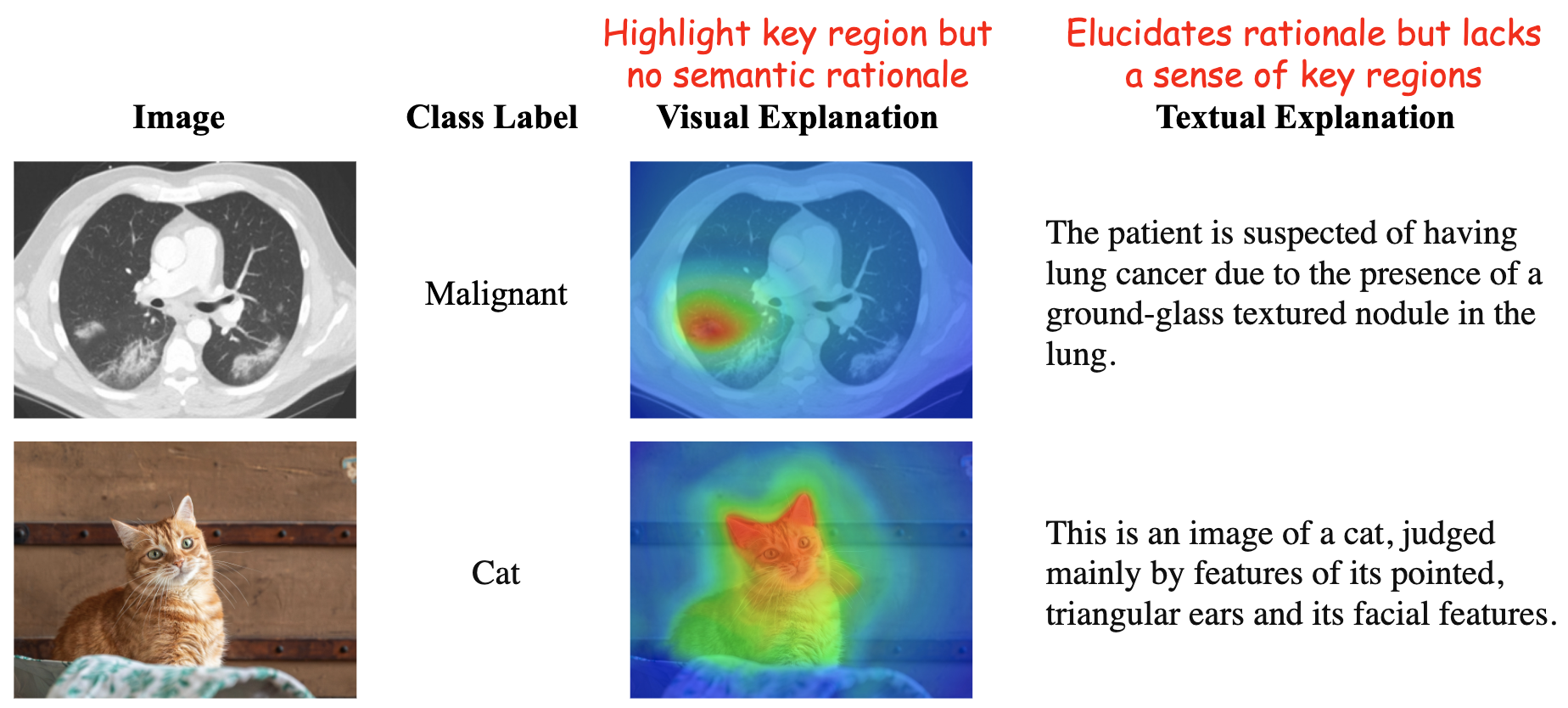}
\caption{Comparison of visual and textual explanations for an image classification task: The visual explanation highlights key regions of interest in the image but lacks a semantic rationale, while the textual explanation provides reasoning behind the decision but lacks spatial context for the key regions.}
\label{fig:example}
\end{figure}

Explaining the decision-making process of Artificial Intelligence (AI) models is essential to addressing their “black box” nature. This need has driven the development of eXplainable AI (XAI) techniques~\cite{du2019techniques, carvalho2019machine, arrieta2020explainable, van2022explainable}, which aim to make model reasoning more transparent and to provide interpretable explanations for AI decisions. However, in the domain of image classification, the current frontier of XAI is limited by two major bottlenecks. Firstly, most XAI methods interpret an image classification model’s decisions through two types of unimodal explanations: visual explanations~\cite{selvaraju2016grad,hendricks2016generating, lundberg2017unified, petsiuk2018rise}, which highlight key areas of the input image, and textual explanations~\cite{hendricks2016generating, cambria2023survey, aminimehr2024tbexplain, speith2022review}, which provide a natural language rationale for the model’s decision-making. However, each unimodal explanation has limitations. For example, as shown in Figure~\ref{fig:example}, in cancer diagnosis, visual explanations highlight key regions of a nodule relevant to the diagnostic decision but may fail to clarify why the nodule is classified as malignant. Textual explanations, on the other hand, can provide rationales, such as identifying ground-glass texture in a lung nodule, but often lack the spatial context provided by the visual explanation. Similiarly, in a cat and dog classification task, visual explanations might highlight key regions like the face and ears of the cat, while textual explanations justify the classification by noting features like the pointed triangular ears.

Secondly, explanations from XAI methods can be inaccurate. For example, visual explanations may highlight incorrect key regions~\cite{geirhos2023don,zhang2022overlooked}, and textual explanations may contain hallucinatory rationales~\cite{biten2022let,ji2023survey,jiang2024hallucination}. To address this issue, Explanation-Guided Learning (EGL)~\cite{tsipras2018robustness, rieger2020interpretations, gao2024going} has emerged as a method to enhance model interpretability and predictive accuracy by refining and aligning the model’s reasoning process with human-understandable explanations. In EGL methods that use visual explanations, existing approaches\cite{zeng2021generating,zhang2023magi,hajialigol2023xai,gu2023essa} employ an explanation loss that compares human-annotated visual explanations with the model’s saliency map, alongside the prediction loss, to guide the model in identifying relevant key regions for its decisions. In contrast, EGL methods that use textual explanations~\cite{li2023symbolic,magister2022teaching,li2022explanations,zhao2023self} employ multitask learning to generate rationales alongside labels in Visual Question Answering tasks to support image classification. However, these textual explanations do not represent the classifier’s true reasoning process but are generated independently by a language model, leaving the model’s internal decision-making as a “black box”~\cite{hendricks2016generating, cambria2023survey, aminimehr2024tbexplain}. Additionally, existing EGL approaches are limited to single-modality explanations, which inherits the aforementioned limitation of using single-modality explanations.

To tackle these issues, this paper paves a new research area called \underline{M}ultimodal \underline{E}xplanation-\underline{G}uided \underline{L}earning (MEGL), which aims to interweave visual and textual explanations' complementary strengthens in visualizing and rationalizing the model decision-making process and improve model's classification performance. However, MEGL requires far more than simply combining visual and textual EGL due to multimodal interdependency and combination: (1) \textit{Interdependence Between Visual and Textual Explanations}: Visual and textual explanations must remain consistent to accurately represent the decision-making process. Textual explanations should incorporate key regions identified by visual explanations, while visual explanations benefit from the semantic context provided by textual rationales. Therefore, effectively coordinating these explanations in EGL remains challenging. (2) \textit{Incompleteness of Explanation Modalities}: In real-world scenarios, acquiring multimodal explanation annotations is often challenging, with some samples lacking one or more modalities due to the resource-intensive nature of generating such annotations. This uneven availability of multimodal explanations leads to heterogeneous supervision signals across the dataset, highlighting the need for a flexible training strategy capable of effectively handling incomplete modalities.

To address the interdependence between visual and textual explanations, we propose a \textit{Saliency-Driven Textual Grounding (SDTG)} approach, which facilitates the transfer of spatial information from visual explanations (saliency map) into textual rationales. SDTG generates textual explanations by combining saliency-driven visual cues with broader contextual information from the full image, ensuring that spatially relevant insights are effectively reflected in the text. Specifically, an input image is processed through a classifier and a post-hoc visual explainer to produce a saliency map that highlights critical regions relevant to the model’s decision. This visual explanation, together with the overall image context, is incorporated into a large language model (LLM) through a carefully constructed prompt that integrates both saliency and image-level features. This process enables the LLM to generate textual explanations that are grounded in spatially relevant information while providing coherent reasoning behind the model’s decisions.

To address the challenge of incomplete explanation modalities, we propose two complementary strategies within our framework. First, our \textit{Saliency-Driven Textual Grounding (SDTG)} approach incorporates \textit{Textual Supervision on Visual Explanations}, leveraging textual rationales to guide the refinement of visual explanations during training. By explicitly transferring and harmonizing information between modalities, SDTG aligns visual explanations more closely with textual perspectives, ensuring that generated explanations are both spatially grounded and mutually informative. Second, for samples without ground truth visual annotations, we introduce a \textit{Visual Explanation Distribution Consistency} loss. This loss aligns the distribution of generated visual explanations with the dataset-level ground-truth distribution, ensuring stable and contextually appropriate visual explanations even in the absence of direct annotations. Together, these methods enable our framework to effectively leverage partial supervision, ensuring meaningful and consistent multimodal explanations across diverse training conditions.

We present two novel datasets, Object-ME and Action-ME, adapted to image classification tasks with multimodal explanation annotations, derived from the VQA-X and ACT-X datasets. These datasets provide both visual and textual explanations, enabling comprehensive evaluation of multimodal explanation-guided learning methods. Extensive experiments conducted on these datasets demonstrate the effectiveness of our proposed MEGL framework. MEGL achieves superior performance over previous image classification and EGL methods, excelling in classification accuracy, visual explainability, and textual explainability.

\section{Related Work}
\subsection{Explanation-Guided Learning}
Many methods have been developed to to leverage explanations to improve the performance of a deep learning model (``Learning with explanations''). CREX~\cite{du2019learning} regularizes deep neural network (DNN) training by enforcing the model to generate local explanations that align with expert-provided rationales, which are subsets of features highlighted as justifications for predictions. CDEP~\cite{rieger2020interpretations} penalize both a model’s prediction and the corresponding explanation.~\cite{zeng2021generating,zhang2023magi,gu2023essa} supervise saliency maps generated by post-hoc explainers to improve image classification performance, while ~\cite{hajialigol2023xai, zhang2016rationale} propose adding supervision on the important tokens in input text for text classification tasks. Some recent work proposes fine-tuning multimodal large language models (MLLMs) using explanations alongside the final answer to enhance their reasoning ability~\cite{li2023symbolic,magister2022teaching,li2022explanations,zhao2023self} for Visual Question Answering tasks for image classification. However, existing work focuses on a single modality of explanation. In this work, we aim to leverage multimodal explanations to enhance the performance of models’ decision-making.

\subsection{Visual and Textual Explanation}
To interpret image classification, visual explanation methods highlight the discriminative regions in the input image, such as through heatmaps, based on gradient or attention maps, like Grad-CAM~\cite{selvaraju2017grad}, Integrated Gradients~\cite{sundararajan2017axiomatic}, and Attention Branch Network (ABN)~\cite{fukui2019attention}. Meanwhile, textual explanations provide a natural language rationale to justify the model’s prediction by a vision-language model as explainer~\cite{hendricks2016generating, aminimehr2024tbexplain, natarajan2024vale}. Additionally, image classification tasks can also be addressed by Visual Question Answering models and generate the natural language rationale~\cite{li2018vqa, wu2018faithful, patro2020robust, rao2021first} with additional explainers.~\cite{park2018multimodal, wu2018faithful} can also generate both visual and textual explanation together. With the development of Multimodal LLMs~\cite{li2023blip, achiam2023gpt, anil2023gemini, liu2024visual}, these models are now capable of generating a sequence that includes both an answer and a textual explanation for a given image~\cite{marasovic2020natural, zhang2023multimodal, hu2024bliva}. In our work, we aim to improve classification performance by correcting both visual and textual explanations while enabling interaction between the two explanation modalities and addressing the challenge of collecting ground truth explanation annotations.

\begin{figure*}[htbp]
  \centering
  \includegraphics[width=0.85\linewidth]{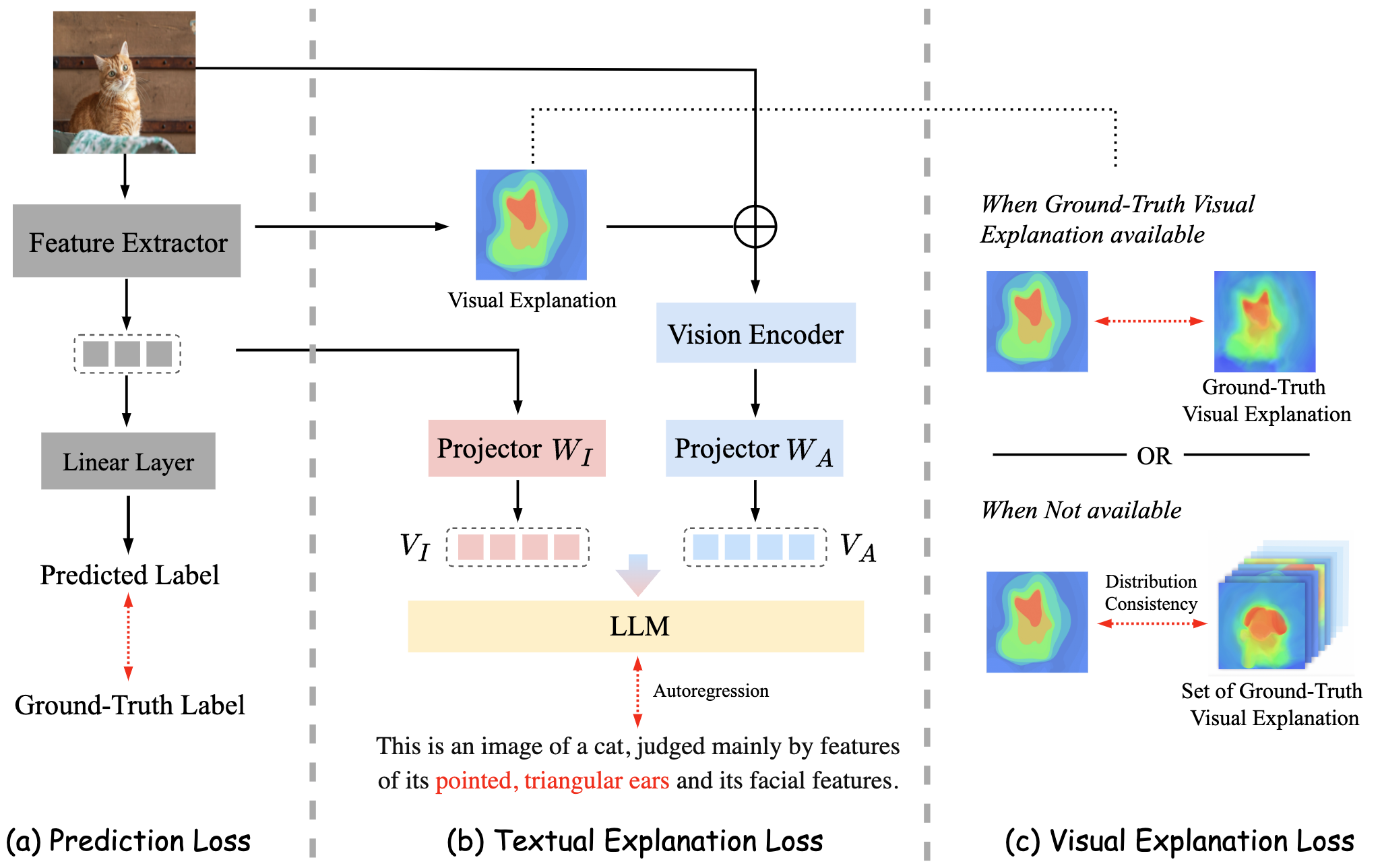}
  \caption{Overview of the MEGL Framework. The framework is jointly trained to optimize prediction accuracy, visual explainability, and textual explainability. (a) illustrates the prediction process, where the input image is processed by the classifier (comprising a feature extractor and a linear layer) to predict the label and extract the image’s visual features. In (c), a saliency map is generated by the visual explanation method as a visual explanation and is used to compute either the visual explanation loss with ground-truth annotations or the distribution consistency loss with the aggregated set of ground-truth visual explanations. In (b), the visual representations of the image and its saliency-based explanation, encoded by a vision encoder, are projected and input into an LLM to generate a textual explanation, supervised by an autoregressive loss. The text in \textcolor{red}{red} (corresponding to the \textcolor{red}{red} regions in the saliency map) showcases how visual cues derived from the saliency map are integrated into the process of generating textual explanations.}
  \label{fig:framework}
\end{figure*}
\section{Methodology}

In this section, we introduce our proposed MEGL framework, beginning with the problem formulation and an overview of framework architecture. We then detail our proposed approaches to facilitating interaction between visual and textual explanations in Section~\ref{sec:text_condition} and address the challenge of incomplete multimodal explanations in Section~\ref{sec:consistency}.

\subsection{Problem Formulation}
\label{problem_formulation}
% During training, the model learns the mapping \( f: (\mathcal{I}) \rightarrow y \), integrating the explanations \( \mathcal{A} \) and \( \mathcal{T} \) to enhance its understanding. At inference time, the model performs the prediction solely based on the input image: \( \hat{y} = f(\mathcal{I}) \), where \( \hat{y} \) represents the predicted label. This framework enables the model to learn from multimodal explanations during training to improve inference accuracy and interpretability, while relying only on the image \( \mathcal{I} \) for prediction during inference.

In multimodal explanation-guided learning for image classification, we aim to enhance a classifier's predictive performance and interpretability by leveraging both visual and textual explanations during training. We define a dataset \( \mathcal{D} \) consisting of triples \( (\mathcal{I}, y, \{\mathcal{A}, \mathcal{T}\}) \), where \( \mathcal{I} \) is the input image, \( y \) is the class label, and \( \{\mathcal{A}, \mathcal{T}\} \) are the associated visual and textual explanations. Our objective is to train a classifier \( f \) that learns the mapping \( f: \mathcal{I} \rightarrow y \) by integrating the explanations \( \mathcal{A} \) and \( \mathcal{T} \) during training. By incorporating these multimodal explanations, the model improves its understanding of the reasoning behind image classifications, leading to enhanced performance and interpretability.

\subsection{Framework Overview}
In this section, we present an overview of the proposed MEGL framework, illustrated in Figure~\ref{fig:framework}. The MEGL framework integrates multimodal learning by incorporating visual and textual explanations as additional supervision to enhance the training of the image classifier.

As illustrated in Figure~\ref{fig:framework}(a), an image classifier \( f \) (e.g., CNN~\cite{lecun2015deep}, ViT~\cite{dosovitskiy2020image}) processes the input image \( \mathcal{I} \) to generate a feature representation by its feature extractor~\cite{lecun2015deep}. This representation is passed through linear layers for classification, producing a logit vector representing class scores. The classifier is trained using a prediction loss 
\[
\mathcal{L}_{\text{pred}}(\phi_f) = \text{CE}(y, \hat{y}), 
\]
where \( \text{CE} \) denotes the cross-entropy loss between the predicted label \( \hat{y} \) and the ground-truth label \( y \), and \( \phi_f \) represents the classifier's parameters.

% As shown in Figure~\ref{fig:framework} (b), a visual explanation method (e.g., Grad-CAM~\cite{selvaraju2017grad}) is applied on the classifier to produce a visual explanation as \( \hat{\mathcal{A}} = f(\mathcal{I})_{\text{visual}} \), highlighting regions of the input \( \mathcal{I} \) that are most relevant to the classification decision. The features of the image from the feature extractor of the classifier, and the features of the visual explanation, encoded by a visual encoder \( E \), are then fed into an LLM to generate a textual explanation that contextualizes the model’s reasoning. The objective function for the textual explanation, 
% \[
% \mathcal{L}_{\text{textual}}(\phi_f, \phi_E, \phi_{\text{LLM}}) = \| \mathcal{T} - \hat{\mathcal{T}} \|_{\text{AR}},
% \] 
% is an autoregressive loss that reduces the discrepancy between the target text \( \mathcal{T} \) and the generated explanation \( \hat{\mathcal{T}} \), where \( \phi_E \) is the parameters of the visual encoder and \( \phi_{\mathcal{LLM}} \) is the parameters of the LLM. Further details on this loss are provided in Section~\ref{sec:text_condition}.

As shown in Figure~\ref{fig:framework} (b), a visual explanation method (e.g., Grad-CAM~\cite{selvaraju2017grad}) is applied to the classifier to generate a visual explanation \( \hat{\mathcal{A}} = f(\mathcal{I})_{\text{visual}} \), highlighting regions of the input \( \mathcal{I} \) that are most relevant to the classification decision. The features of the image, extracted by the classifier, and the visual explanation, encoded by a visual encoder \( E \), are combined and fed into an LLM to generate a textual explanation. This textual explanation process is supervised by an autoregressive loss 
\[
\mathcal{L}_{\text{textual}}(\phi_f, \phi_E, \phi_{\text{LLM}}) = \| \mathcal{T} - \hat{\mathcal{T}} \|_{\text{AR}},
\]
which encourages the generated explanation \( \hat{\mathcal{T}} \) to align with the target rationale \( \mathcal{T} \). Details are provided in Section~\ref{sec:text_condition}.

As shown in Figure~\ref{fig:framework} (c), with the generated visual explanation $\hat{\mathcal{A}}$ is supervised by the ground-truth visual explanation \( \mathcal{A} \) as
\begin{equation}
\mathcal{L}_{\text{visual}}(\phi_f) = \| \hat{\mathcal{A}} - \mathcal{A} \|_1,
\label{eq:visual_exp_sup_v0}
\end{equation}
which minimizes the \( L_1 \) distance between the generated visual explanation and the ground-truth visual explanation. For samples lacking ground truth visual explanations, we introduce a visual explanation distribution consistency loss, defined as 
\[
\mathcal{L}_{\text{dc}}(\phi_f) = \mathbb{E}\big[ \| \hat{\mathcal{A}} - \textbf{A} \| \big],
\]
where \( \textbf{A} \) denotes the set of ground truth visual explanations across the dataset. This consistency loss ensures that the generated saliency maps align with the overall distribution of ground truth annotations. Depending on the availability of visual annotations, either \( \mathcal{L}_{\text{visual}} \) or \( \mathcal{L}_{\text{dc}} \) is applied. Further details are provided in Section~\ref{sec:consistency}.

The final objective function is designed to jointly optimize classification accuracy, visual explainability, and textual explainability, and is formulated as follows:
\begin{align*}
\mathcal{L} = \, &\mathcal{L}_{\text{pred}}(\phi_f) \\
&+ \lambda_{\text{textual}} \mathcal{L}_{\text{textual}}(\phi_f, \phi_E, \phi_{\text{LLM}}) \\
&+ \lambda_{\text{visual}} \left( \mathbb{I}_{\text{V}} \mathcal{L}_{\text{visual}}(\phi_f) + (1 - \mathbb{I}_{\text{V}}) \mathcal{L}_{\text{dc}}(\phi_f) \right),
\end{align*}
where \( \mathbb{I}_{\text{V}} \) is an indicator function set to 1 when visual annotations are available, and 0 otherwise. The hyperparameters \( \lambda_{\text{visual}} \) and \( \lambda_{\text{textual}} \) control the balance between the prediction loss, visual explanation loss, and textual explanation loss.

\subsection{Facilitating Interaction in Multimodal Explanations}
\label{sec:text_condition}
In this section, we present the \textit{Saliency-Driven Textual Grounding} (SDTG) method, designed to facilitate interaction between visual and textual explanations. Unlike traditional end-to-end image-to-text generation approaches~\cite{hendricks2016generating, cambria2023survey, aminimehr2024tbexplain}, our SDTG method generates textual explanations by adding visual explanations as input, leveraging spatial information to ground the semantic rationale in the identified key regions.

To be specific, firstly, as shown in Figure~\ref{fig:framework} (a) and (b), for a given input image \( \mathcal{I} \), we generate a visual explanation \( \hat{\mathcal{A}} \) with a visual explanation method, such as Grad-CAM, as \( \hat{\mathcal{A}} = f(\mathcal{I})_{\text{saliency}} \). The image and visual explanation are then processed through two encoding paths to obtain the visual representation. In the first path, we obtain the vision representation of the original image \( \mathcal{I} \) by the feature extractor of classifier$f$ as $f(\cdot)_{\text{feature}}$. In the second path, we apply the visual explanation to the image, yielding a modified image as \( \hat{\mathcal{I}} = \mathcal{I} \odot \hat{\mathcal{A}} \), which is then encoded with the pre-trained CLIP vision encoder~\cite{radford2021learning}, denoted \(\text{CLIP} \), to capture the visual representation of the visual explanation. 

We apply two trainable projection matrices, \( W_{\text{I}} \) and \( W_{\text{A}} \), to map the visual representations of the original image and its corresponding saliency-based visual explanation, \( \mathbf{Z}_{\text{I}} \) and \( \mathbf{Z}_{\text{A}} \), into  language embedding tokens \( \mathbf{V}_{\text{I}} \) and \( \mathbf{V}_{\text{A}} \), which have the same dimensionality as the language model’s word embeddings:
\[
\mathbf{V}_{\text{I}} = W_{\text{I}} \cdot \mathbf{Z}_{\text{I}}, \quad \text{where } \mathbf{Z}_{\text{I}} = f(\mathcal{I})_{\text{feature}},
\]
\[
\mathbf{V}_{\text{A}} = W_{\text{A}} \cdot \mathbf{Z}_{\text{A}}, \quad \text{where } \mathbf{Z}_{\text{A}} = \text{CLIP}(\hat{\mathcal{I}}).
\]
Then, we feed both \( \mathbf{V}_{\text{I}} \) and \( \mathbf{V}_{\text{A}} \) into an LLM to generate the textual explanation guided by the saliency map. Specifically, the LLM, denoted as \(\text{LLM}(\cdot)\), takes the combined visual representation \( \mathbf{V} = \mathbf{V}_{\text{A}} + \mathbf{V}_{\text{I}} \) as input and generates a sequence of tokens \( \mathcal{T} = (t_1, t_2, \ldots, t_n) \) representing the textual explanation:
\[
\mathcal{T} = \text{LLM}(\mathbf{V}), \quad \text{where } \mathbf{V} = \mathbf{V}_{\text{A}} + \mathbf{V}_{\text{I}}.
\]
The textual explanation loss \( \mathcal{L}_{\text{Textual}} \) is defined as an autoregressive loss, ensuring that the generated explanation \( \mathcal{T} \) matches the ground truth \( \mathcal{T}^* \). Using a negative log-likelihood formulation, we have:
\begin{equation}
\mathcal{L}_{\text{Textual}} = - \sum_{i=1}^{n} \log P(t_i = t_i^* \mid \mathbf{V}, t_{<i}),
\label{eq:text_exp_sup}
\end{equation}
where \( P(t_i = t_i^* \mid \mathbf{V}, t_{<i}) \) is the model's probability of generating the correct token \( t_i \) at each step, conditioned on the combined visual representation \( \mathbf{V} \) and the sequence of previously generated tokens \( t_{<i} \). This method achieves a well-grounded interaction between visual and textual explanations, leveraging spatial cues to enhance the semantic coherence of generated textual rationales.

\subsection{Handling Multimodal Explanation Incompleteness}
\label{sec:consistency}
In this section, we present our approach to addressing the incompleteness of multimodal explanations. Specifically, we tackle this challenge through two strategies: \textit{Textual Supervision on Visual Explanations} and \textit{Visual Explanation Distribution Consistency}.

\subsubsection{Textual Supervision on Visual Explanations}
Firstly, in our SDTG method, \textit{Textual Supervision on Visual Explanations} is achieved by guiding the textual explanation generation process with the visual explanation produced by the classifier's visual explainer. This interaction encourages the visual explainer to iteratively refine its output, \( \hat{\mathcal{A}} \), based on feedback from the textual rationale, thereby fostering alignment between visual and textual explanations. Even in the absence of ground truth visual annotations, this approach enables effective supervision through the textual modality, allowing the model to leverage textual guidance during training.

The textual explanation generation process leverages both the input image representation \( \mathbf{V}_{\text{I}} \) and the visual explanation representation \( \mathbf{V}_{\text{A}} \). Minimizing the textual explanation loss \( \mathcal{L}_{\text{Textual}} \) encourages consistency between the generated saliency map \( \hat{\mathcal{A}} \) and the textual rationale, promoting alignment as shown in:
\[
\hat{\mathcal{A}} = \arg \min_{\hat{\mathcal{A}}} \mathcal{L}_{\text{Textual}}(\mathbf{V}_{\text{I}}, \mathbf{V}_{\text{A}})
\]
Through this alignment, \( \mathcal{L}_{\text{Textual}} \) provides semantic grounding, ensuring that \( \hat{\mathcal{A}} \) reflects the rationale expressed in the textual explanation, enabling the visual explanation to be iteratively refined according to the textual perspective.

\subsubsection{Visual Explanation Distribution Consistency}
We propose a \textit{Visual Explanation Distribution Consistency} method to generate meaningful visual explanations even for samples without ground truth annotations, thus enabling more robust training across the entire dataset. Specifically, we introduce a consistency loss that aligns the distribution of generated visual explanations with the aggregated distribution of available ground truth visual explanations. This method leverages the relationship between annotated and unannotated samples, providing supervision at the distributional level in the absence of direct paired supervision.

For samples without ground truth visual explanation annotations, we generate a saliency map, \( \hat{\mathcal{A}} \). For annotated samples, we construct an aggregated target distribution \( \bar{\mathcal{A}} \) by averaging their normalized ground truth saliency maps, capturing the typical pattern of visual explanations across annotated data:
\[
\bar{\mathcal{A}}(i, j) = \frac{1}{n} \sum_{k=1}^{n} \mathcal{A}_{k}(i, j),
\]
where \( n \) represents the number of annotated samples. This aggregated distribution \( \bar{\mathcal{A}} \) serves as a reference pattern for unannotated samples, reflecting the average distribution of visual explanations within the dataset.

To enforce consistency, we define the consistency loss \( \mathcal{L}_{\text{ds}} \) as the Kullback-Leibler divergence between the generated saliency map \( \hat{\mathcal{A}} \) and the aggregated distribution \( \bar{\mathcal{A}} \):
\[
\mathcal{L}_{\text{dc}}(\hat{\mathcal{A}}, \bar{\mathcal{A}}) = D_{\text{KL}}(\hat{\mathcal{A}} \parallel \bar{\mathcal{A}}),
\]
which encourages the model to align generated visual explanations with the distribution patterns of the annotated ground truth, promoting consistent visual explanations across the dataset.

With the proposed consistency loss \( \mathcal{L}_{\text{ds}} \), the visual explanation loss as in Equation~\ref{eq:visual_exp_sup_v0} can be extended. For samples with ground truth visual explanations, we apply the direct explanation loss \( \mathcal{L}_{\text{visual}}(\hat{\mathcal{A}}, \mathcal{A}) \), where \( \mathcal{A} \) represents the ground truth saliency map. For samples without ground truth, we instead apply the distribution consistency loss \( \mathcal{L}_{\text{dc}}(\hat{\mathcal{A}}, \bar{\mathcal{A}}) \). The combined objective is expressed as:
\[
\mathcal{L}_{\text{Visual}} = \mathbb{I}_{\text{V}} \mathcal{L}_{\text{visual}}(\hat{\mathcal{A}}, \mathcal{A}) + (1 - \mathbb{I}_{\text{V}}) \mathcal{L}_{\text{dc}}(\hat{\mathcal{A}}, \bar{\mathcal{A}})
\]
where \( \mathbb{I}_{\text{V}} \) is an indicator function that equals 1 when ground truth annotations are available and 0 otherwise. This combined objective encourages the model to generate consistent and meaningful visual explanations across all samples, leveraging both direct supervision and distributional alignment for interpretability even in partially annotated datasets.

% You must include your signed IEEE copyright release form when you submit your finished paper.
% We MUST have this form before your paper can be published in the proceedings.

% Please direct any questions to the production editor in charge of these proceedings at the IEEE Computer Society Press:
% \url{https://www.computer.org/about/contact}.
\section{Experiment}

\subsection{Dataset} 
We experiment with our constructed two datasets for image classification with both visual and textual explanation annotations, derived from two VQA with visual and tecxtual explanations datasets: Visual Question Answering Explanation (VQA-X) and Activity Explanation (ACT-X)~\cite{park2018multimodal}. We extract samples that can be converted into a classification task from the VQA-X dataset and construct the Object-ME dataset for object classification. Similarly, the Action-ME dataset is derived from the ACT-X for action classification. For the two constructed datasets, each image sample includes a class label and a corresponding textual explanation to justify the class label. Additionally, a subset of the samples contains visual explanations to further support the justification of the class label. Table~\ref{tab:dataset} summarizes the statistics of the two proposed datasets, detailing the total number of samples, the number of samples with visual explanations, the number of samples with textual explanations, and the total number of class labels.

\begin{table}[htbp]
\centering
\resizebox{\linewidth}{!}{
\begin{tabular}{lcccc}
\toprule
Dataset & Total Samples & Textual Exp. & Visual Exp. & \# Classes\\ 
\midrule
Object-ME  & 4,790  & 4,790  & 402  & 40 \\
Action-ME  & 11,511 & 11,511 & 1,185 & 127 \\ 
\bottomrule
\end{tabular}
}
\caption{Dataset Statistics for Object-ME and Action-ME.}
\label{tab:dataset}
\end{table}

\subsection{Evaluation Metrics}
To comprehensively evaluate the effectiveness of our proposed MEGL framework, we conduct assessments across three key dimensions: classification performance, visual explainability, and textual explainability. For classification performance, we employ standard metrics including Accuracy, Precision, Recall, and F1-score. For visual explainability, we use the mean Intersection-over-Union (mIoU) to quantify the overlap between the generated visual explanations and ground truth visual annotations, divided by the total area covered by the union of the two. For textual explainability, we assess both the quality and the faithfulness of the generated textual explanation. Specifically, the quality of generated textual explanations is measured using established metrics: BLEU-4~\cite{papineni2002bleu}, METEOR~\cite{banerjee2005meteor}, ROUGE-L~\cite{lin2004rouge}, CIDEr~\cite{vedantam2015cider}, and SPICE~\cite{anderson2016spice}. To evaluate the faithfulness~\cite{zhou2023analyzing,hashem2024generating} of the generated textual explanation, we utilize the CLIPScore~\cite{hessel2021clipscore} to measure text-image alignment. In addition, we evaluate the efficiency of the frameworks by analyzing their number of parameters, latency, and frames per second (FPS).

\begin{table*}[htbp]
\centering
\resizebox{\linewidth}{!}{%
\begin{tabular}{llcccccccccc}
\toprule
\multirow{2}{*}{Backbone} & \multirow{2}{*}{Method} & \multicolumn{5}{c}{Object-ME} & \multicolumn{5}{c}{Action-ME} \\
\cmidrule(lr){3-7} \cmidrule(lr){8-12}
& & Accuracy & Precision & Recall & F1 Score & mIoU & Accuracy & Precision & Recall & F1 Score & mIoU \\
\midrule
LLaVA~\cite{liu2024visual} & - & 0.8220 & 0.5702 & 0.5550 & 0.5568 & - & 0.8595 & 0.6436 & 0.6124 & 0.6233 & -  \\
LLaVA & Fine-Tune-CoT~\cite{ho2022large} & 0.8233 & 0.5448 & 0.6003 & 0.5643 & - & 0.8789& 0.7052& 0.6182& 0.6566& - \\
\midrule
\multirow{6}{*}{ResNet18}     
& - & 0.7265 & 0.5294 & 0.5432 & 0.5192 & 0.3391 & \underline{0.7973} & \underline{0.7807} & \textbf{0.8307} & \textbf{0.7952} & 0.3961 \\ 
& CDEP~\cite{rieger2020interpretations} & 0.7119 & 0.5320 & 0.5925 & 0.5400 & \underline{0.3703} & 0.7764 & 0.7514 & 0.8147 & 0.7651 & 0.4168 \\ 
& HAICS~\cite{shen2021human} & 0.7203 & 0.5208 & 0.5507 & 0.5181 & 0.3692 & 0.7649 & 0.7379 & 0.8046 & 0.7493 & 0.4142\\
& RES-G~\cite{gao2022res} & 0.7171 & 0.5402 & 0.5944 & 0.5439 & 0.3633 & 0.7799 & 0.7579 & 0.8094 & 0.7694 & \textbf{0.4213}\\
& RES-L~\cite{gao2022res} & \underline{0.7307} & \textbf{0.5704} & \underline{0.6294} & \underline{0.5677} & 0.3688 & 0.7892 & 0.7696 & 0.8169 & 0.7813 & 0.4045\\
& MEGL \cellcolor{gray!15} & \cellcolor{gray!15}\textbf{0.7413} & \cellcolor{gray!15}\underline{0.5689} & \cellcolor{gray!15}\textbf{0.6595} & \cellcolor{gray!15}\textbf{0.5800} & \cellcolor{gray!15}\textbf{0.3893} & \cellcolor{gray!15}\textbf{0.8025}& \cellcolor{gray!15}\textbf{0.7855} & \cellcolor{gray!15}\underline{0.8246} & \cellcolor{gray!15}\underline{0.7937} & \cellcolor{gray!15}\underline{0.4195}\\
% & MEGL-Q \cellcolor{gray!15} & \cellcolor{gray!15} & \cellcolor{gray!15} & \cellcolor{gray!15} & \cellcolor{gray!15} & \cellcolor{gray!15} & \cellcolor{gray!15} & \cellcolor{gray!15} & \cellcolor{gray!15} & \cellcolor{gray!15} & \cellcolor{gray!15} \\
\bottomrule
\multirow{6}{*}{ViT-B/16} 
& - & 0.7858 & 0.6460 & 0.6456 & 0.6352 & 0.3323 & \underline{0.8854} & 0.8771 & 0.8981 & 0.8803 & 0.3351\\
& CDEP~\cite{rieger2020interpretations} & 0.8150 & \underline{0.7014} & 0.7013 & \underline{0.6911} & \textbf{0.3556} & 0.8836 & 0.8771 & 0.8880 & 0.8770 & 0.3582\\
& HAICS~\cite{shen2021human} & 0.8178 & 0.6722 & 0.6813 & 0.6642 & 0.3443 & \underline{0.8854} & \underline{0.8784} & \underline{0.8924} & \underline{0.8807} & 0.3557\\
& RES-G~\cite{gao2022res} & 0.8164& 0.6864& 0.7094& 0.6833& 0.3441 & 0.8796 & 0.8726 & 0.8851 & 0.8738 & 0.3604 \\
& RES-L~\cite{gao2022res} & 0.8206 & 0.6870 & \underline{0.7296} & 0.6850 & 0.3401& 0.8761 & 0.8712 & 0.8806 & 0.8677 & 0.3639 \\
% & Fine-Tune-CoT~\cite{ho2022large} & 0.8233 & 0.5448 & 0.6003 & 0.5643 & - & 0.8789& 0.7052& 0.6182& 0.6566& - \\
& MEGL \cellcolor{gray!15} & \cellcolor{gray!15}\textbf{0.8317}& \cellcolor{gray!15}\textbf{0.7037}& \cellcolor{gray!15}\textbf{0.7485}& \cellcolor{gray!15}\textbf{0.7036}& \cellcolor{gray!15}\underline{0.3521}& \cellcolor{gray!15}\textbf{0.8981}& \cellcolor{gray!15}\textbf{0.8897}& \cellcolor{gray!15}\textbf{0.9024}& \cellcolor{gray!15}\textbf{0.8921}& \cellcolor{gray!15}\textbf{0.3681}\\
% & MEGL-Q \cellcolor{gray!15} & \cellcolor{gray!15} & \cellcolor{gray!15} & \cellcolor{gray!15} & \cellcolor{gray!15} & \cellcolor{gray!15} & \cellcolor{gray!15} & \cellcolor{gray!15} & \cellcolor{gray!15} & \cellcolor{gray!15} & \cellcolor{gray!15} \\
\bottomrule
\end{tabular}%
}
\caption{\textbf{Comparison with SOTA EGL Methods: Classification and Visual Explainablity}. Macro-average classification metrics and mIoU are reported to to compare classification performance and visual explainablity across datasets. For textual EGL methods, only classification performance is reported. The best performing values of each backbone are shown in bold, while the second-best values of each backbone are marked with underline. Our MEGL models outperform corresponding baselines and MEGL-ViT-B/16 outperforms all SOTA baseline models}
% Our MEGL-ViT-B/16 outperforms all SOTA baseline models.
\label{tab:accuracy_miou_comparison}
\end{table*}

\begin{table*}[htbp]
\centering
\resizebox{\linewidth}{!}{
\begin{tabular}{lccccccccccccc}
\toprule
& \multicolumn{6}{c}{Object-ME} & \multicolumn{6}{c}{Action-ME} \\
\cmidrule(lr){2-7} \cmidrule(lr){8-13}
Method & CLIPScore & B4 & M & R & S & C & CLIPScore & B4 & M & R & S & C \\
\midrule 										
LLaVA & 0.6512 & 0.6772 & 0.5892 & 0.6973 & 0.6975 & 6.9684 & 0.6437 & 0.6005 & 0.5212 & 0.5212 & 0.6319 & 6.1986 \\
LLaVA-Fine-tune-CoT & 0.6550 & 0.7032 & \underline{0.6486} & 0.7168 & 0.7167 & \underline{7.5177} & 0.6510 & 0.6448 & \underline{0.5584} & 0.6616 & 0.6650 & \underline{6.5285} \\
\midrule 					
\rowcolor{gray!15} MEGL-ResNet18 & \underline{0.6613} & \underline{0.7353} & 0.6335 & \underline{0.7474} & \underline{0.7473} & 7.4691 & \underline{0.6529} & \underline{0.6570} & 0.5535 & \underline{0.6697} & \underline{0.6731} & 6.6096 \\ 		
\rowcolor{gray!15} MEGL-ViT-B/16 & \textbf{0.6698} & \textbf{0.7817} & \textbf{0.6701} & \textbf{0.7863} & \textbf{0.7863}  & \textbf{7.8585}  & \textbf{0.6586} & \textbf{0.6851} & \textbf{0.5775} & \textbf{0.6946} & \textbf{0.6980} & \textbf{6.8585} \\
\bottomrule
\end{tabular}
}
\caption{\textbf{Comparison with SOTA EGL Methods: Textual Explanation Quality}. Comparison of MEGL with Fine-tune-CoT across different language models and datasets are reported. The B4, M, R, S, and C are short for BLEU-4, METEOR, ROUGE-L, SPICE, CIDEr, respectively. The best performing values are shown in bold, while the second-best values are marked with underline. Our MEGL-ViT-B/16 outperforms all SOTA baseline models.}
\label{tab:textual_explanation_evaluation}
\end{table*}

\subsection{Comparison Methods}
To validate the effectiveness of our proposed framework, we conduct comprehensive evalauations of our MEGL framework against various baseline models and state-of-the-art approaches across the three key dimensions mentioned above: classification performance, visual explainability, and textual explainability. 

We take into account 3 categories of models: traditional vision models, multimodal large language models (MLLMs), and state-of-the-art EGL frameworks. The vision baselines include Convolutional Neural Networks (CNNs) and Vision Transformer models (ViTs), represented by ResNet18 ~\cite{he2016deep}  and ViT-B/16~\cite{dosovitskiy2020image} respectively. Additionally, we evaluate against state-of-the-art MLLMs such as LLaVA ~\cite{liu2024visual}, which are fine-tuned to perform image classification as a visual question-answering task. Moreover, for the state-of-the-art EGL frameworks, we evaluate both visual EGL and textual EGL architectures. We evaluate established visual EGL frameworks including CDEP~\cite{rieger2020interpretations}, HAICS~\cite{shen2021human}, RES-G, and RES-L~\cite{gao2022res} which focus on optimizing spatial regions used by the model to enhance model interpretability. For textual-based EGL, we evaluate against Fine-tune-CoT~\cite{ho2022large} , which adopts a VQA-style approach similar to LLaVA and fine-tunes large language models to simultaneously generate predictions and textual explanations from visual inputs. 

For classification performance, we benchmark against all three categories. For visual explainability, we evaluate our proposed MEGL framework against the traditional vision models and the state-of-the-art visual EGL frameworks. For textual explainability, we evaluate against the state-of-the-art textual EGL frameworks.

\subsection{Implementation Details}
For MEGL, we utilize ResNet18 and ViT-B/16 as backbone classifiers. The pre-trained vision encoder is CLIP-ViT-L-14, while the LLM is Vicuna v1.5~\cite{zheng2023judging}. Fine-tuning is performed using the LLaVA-1.5-7B checkpoint. Fine-tuning of the MLLMs is based on the LLaVA framework, specifically leveraging the LLaVA-1.5-7B model. For generating visual explanations from image classification models, we use Grad-CAM~\cite{selvaraju2017grad}. 

All experiments are conducted on four NVIDIA A6000 GPUs. During the fine-tuning of both MLLMs and MEGL, the vision tower is frozen, and adaptation is carried out using LoRA~\cite{hu2021lora}.

\subsection{Main Results} 

\noindent{\bf Comparisons of Classification Performance} are presented in Table~\ref{tab:accuracy_miou_comparison}. Our MEGL methods outperform the corresponding baseline models in image classification performance on both Object-ME and Action-ME datasets, and exhibit substantial improvement compared with existing visual EGL methods. Notably, the MEGL with ViT-B/16 backbone classifier (MEGL-ViT-B/16) achieved the best classification performance among all models in the evaluation on both datasets. 

We found that visual EGL methods achieved marked improvements across metrics beyond accuracy. However, this may come at the cost of a slight reduction in accuracy scores, which might be caused by the additional saliency-guided learning procedure. The phenomenon is more significant in models with ResNet18 backbone, which could possibly be explained by relatively simpler architecture compared with ViT-B/16. 

We also found that MLLMs and MLLM-based textual EGL frameworks exhibit excellent classification performance in terms of Accuracy. However, they demonstrate relatively poor performance across Precision, Recall, and F1-score metrics, particularly on Action-ME dataset. This is possibly due to the fact that decision-making of such MLLMs could be unstable in edge cases in image classification. 

\noindent{\bf Comparisons of Visual Explainability} are also presented in Table~\ref{tab:accuracy_miou_comparison}. Our methods and EGL frameworks demonstrate substantial improvements in the quality of generated visual explanations, with particularly pronounced enhancements observed in the performance of ResNet18-based models. Notably, the mIoU metrics achieved on the Action-ME dataset significantly exceeded those on Object-ME under the setting of ResNet18-based models. Such disparity can be attributed to the differential distribution of visual explanation samples between the datasets, with Action-ME containing approximately three times the number of annotated samples compared to Object-ME. These findings underscore the critical importance of visual explanation distribution consistency, as it enables optimal utilization of existing visual explanations and helps address the fundamental challenge of annotation scarcity.

Our MEGL framework achieved significant improvements across all other performance metrics, indicating comprehensive enhancement in the overall classification capabilities and visual explainability. Of particular note, MEGL-ViT-B/16 model achieves the best performance across all evaluation metrics on both datasets in classification tasks and also exhibit top visual explainability, further validating the effectiveness of our proposed framework.

\noindent{\bf Comparison of Textual Explainability} is shown in Table \ref{tab:textual_explanation_evaluation}. Our MEGL-ViT-B/16 model consistently outperform both fine-tuned LLaVA and the LLaVA-Fine-tune-CoT in terms of CLIPScore, demonstrating superior capabilities in generating faithful textual explanations. This improvement in CLIPScore indicates enhanced text-image alignment and suggests that our models achieve better semantic consistency between visual inputs and textual outputs. This performance advantage manifests in the generation of more faithful textual explanations that exhibit stronger semantic alignment with visual inputs. 

In addition, MEGL-ViT-B/16 also generates texutal explanations of higher quality than the  LLaVA-Fine-tune-CoT. We observe that MEGL with ResNet18 backbone classifier (MEGL-ResNet18) demonstrates superior performance over LLaVA-Fine-tune-CoT on the Object-ME dataset, while both models achieve comparable performance levels on Action-ME. This could be possibly accounted by the larger size of Action-ME. It is worth noting that the high scores achieved across various language metrics may be attributed to the relatively templated nature of textual explanations in our datasets. Large language models, after fine-tuning, can readily generate responses that conform to these templates, potentially leading to inflated metric scores. 

\subsection{Ablation study}
Ablation study is conducted on our proposed MEGL framework to validate its effectiveness. Components responsible for visual explanations and textual explanations are removed respectively. 

The results of ablation study is shown in Table \ref{tab:ablation}. It is clear that for MEGL models with backbone of ResNet18 and ViT-B/16, all the components created positive impact on the performance. In addition, the positive effect of textual explanation is more significant compared with that of visual explanation and visual explanation distribution consistency.

\begin{table}[htbp]
\resizebox{\linewidth}{!}{
\begin{tabular}{llcc}
\toprule
Backbone                  & Modules                                     & Accuracy & Performance Change \\ \midrule
\multirow{5}{*}{Resnet18} & MEGL & 0.7413   & -                  \\
                          & Visual+Text                                 & 0.7399   & -0.14\%            \\
                          & Text                                        & 0.7385   & -0.28\%            \\
                          & Visual                                      & 0.7371   & -0.42\%            \\
                          & -                                           & 0.7265   & -1.48\%            \\ \midrule
\multirow{5}{*}{ViT-B/16} & MEGL & 0.8317   & -                  \\
                          & Visual+Text                                 & 0.8234   & -0.83\%            \\
                          & Text                                        & 0.8220   & -0.97\%            \\
                          & Visual                                      & 0.8206   & -1.11\%            \\
                          & -                                           & 0.7858   & -4.59\%            \\ \bottomrule
\end{tabular}
}
% \caption{Evaluation of Textual Explanation Quality: Comparison of MEGL with Fine-tune-CoT across Different Language Models and Datasets. The B4, M, R, S, and C are short for BLEU-4, METEOR, ROUGE-L, SPICE, CIDEr, respectively.}
\caption{\textbf{Ablation study}. Performed on Object-ME dataset with different combinations of components. Visual, Text and Consistency stands for visual explanation modules, textual explanation modules and visual explanation distribution consistency.}
\label{tab:ablation}
\end{table}

\subsection{Efficiency Analysis}

Although MEGL models achieved significant improvement upon the corresponding baseline models, their enhancement upon LLaVA based models, especially LLaVA-Fine-tune-CoT, seems marginal. However, it should be noticed that MEGL models are much smaller and of higher efficiency when deployed for image classification tasks.

To demonstrate the advantages in efficiency our proposed MEGL framework, we conduct comprehensive analyses on model size and computational costs of MEGL models and LLaVA-Fine-tune-CoT on image classification tasks and present the results in Table \ref{tab:efficiency_latency_fps}. We can see that the FPS of MEGL models are significantly higher than LLaVA-Fine-tune-CoT, suggesting higher inference speed and higher efficiency. The discrepancy translates to a 30.1$\times$ speedup for MEGL-ViT-16/B in FPS with improved classification performance, illustrating the effectiveness and efficiency of our MEGL framework. 

\begin{table}[htbp]
\resizebox{\linewidth}{!}{
\begin{tabular}{lcccc}
\toprule
Model         & Parameters & Latency   & FPS  & Speedup  \\ \midrule
MEGL-ResNet18 & 11M        & 8.52 ms   & 117.41 & 41.78$\times$ \\
MEGL-ViT-16/B & 85M        & 11.82 ms  & 84.61 & 30.11$\times$  \\
LLaVA-Fine-tune-CoT     & 7063M      & 356.47 ms & 2.81 & -  \\ \bottomrule
\end{tabular}
}
\caption{\textbf{Efficiency Analysis}. All models are tested with BF16 precision and batch size of 1 on an NVIDIA L40S. The latency and FPS in this table are measured without post-processing or decoding. Speedup is calculated based on the FPS of LLaVA-Fine-tune-CoT. }
\label{tab:efficiency_latency_fps}
\end{table}

% You must include your signed IEEE copyright release form when you submit your finished paper.
% We MUST have this form before your paper can be published in the proceedings.

% Please direct any questions to the production editor in charge of these proceedings at the IEEE Computer Society Press:
% \url{https://www.computer.org/about/contact}.
\section{Conclusion}
In this paper, we introduced the Multimodal Explanation-Guided Learning (MEGL) framework, designed to integrate multimodal explanations and enhance classification performance. MEGL incorporates Saliency-Driven Textual Grounding (SDTG), which facilitates interaction between multimodal explanations, ensuring alignment and mutual consistency while also enabling Textual Supervision on Visual Explanations, where textual rationales refine visual explanations during training. Additionally, the Visual Explanation Distribution Consistency loss tackles the challenge of incomplete visual annotations by generating robust visual explanations even for unannotated samples. Extensive experiments on two newly proposed datasets, Object-ME and Action-ME, demonstrate that MEGL outperforms existing methods in classification accuracy, visual explainability, as well as textual explainability. By effectively leveraging multimodal explanations, MEGL advances both the interpretability and predictive performance of AI systems.

{
    \small
    \bibliographystyle{ieeenat_fullname}
    \bibliography{main}
}


\end{document}